\documentclass[a4paper]{llncs}

\usepackage{etex}
\usepackage{times}
\usepackage{graphicx}
\usepackage{float,latexsym,url}
\usepackage[ruled,vlined]{algorithm2e}
\usepackage{pstricks-add}
\usepackage{pgf}
\usepackage[utf8]{inputenc}
\usepackage{colortbl}

\usepackage{amsmath}
\usepackage{amssymb}
\usepackage{latexsym,xspace}
\usepackage{graphics,graphicx,float}
\usepackage{times,psfrag}
\usepackage{pstricks-add}
\usepackage{rotate}
\usepackage{epsfig}
\usepackage{subfigure}
\usepackage{textcomp}
\usepackage{tikz}
\usetikzlibrary{arrows,shapes,snakes,automata,backgrounds,petri,shadows}

\usepackage{latexsym}

\usepackage{proof}

\usepackage{url}
\urldef{\mailsa}\path|{alfred.hofmann, ursula.barth, ingrid.haas, frank.holzwarth,|
\urldef{\mailsb}\path|anna.kramer, leonie.kunz, christine.reiss, nicole.sator,|
\urldef{\mailsc}\path|erika.siebert-cole, peter.strasser, lncs}@springer.com|

\newcommand{\nomore}[1]{}

\begin{document}

\mainmatter  

\title{On SAT Models Enumeration in Itemset Mining}
\author{Said Jabbour and Lakhdar Sais and Yakoub Salhi}

\institute{
CRIL-CNRS Univ. Artois\\
F-62307, Lens \\
France \\
\{jabbour, sais, salhi\}@univ-orleans.fr
}

\maketitle

\begin{abstract}
Frequent itemset mining is an essential part of data analysis and data mining.
Recent works propose interesting SAT-based encodings for the problem of discovering 
frequent itemsets. Our aim in this work is to define strategies for adapting 
SAT solvers to such encodings in order to improve models enumeration.
In this context, we deeply study the effects of restart, branching heuristics and clauses learning.
We then conduct an experimental evaluation on SAT-Based itemset mining instances to show how SAT solvers can be adapted to obtain an efficient SAT model enumerator. 
\end{abstract}
 
\section{Introduction}
\label{sect:introduction}

Frequent itemset mining is a keystone in several data analysis and data mining tasks. Since the first article of Agrawal \cite{agrawal93} on association rules and itemset mining, the huge number of works, challenges, datasets and projects show the actual interest in this problem (see \cite{tiwari2010survey} for a  survey of works addressing this problem). 
In~ \cite{Raedt08}, De Raedt  et al initiate a research  trend on constraint programming and data mining. The authors proposed a framework 
for itemset mining offering a declarative and flexible representation with several generic and efficient CP solving techniques.
Encouraged by the promising results of this framework, several contributions addressed other data mining problems using either constraint programming (CP) or 
 propositional satisfiability (SAT) (e.g. \cite{Raedt08,GunsNR11,Lynce12Itemset,CoqueryJSS12,JabbourSS13topk,JabbourSS13seq}).
 In this work, we focus on the SAT-based encoding of itemset mining problems \cite{JabbourSS13topk}. In this new SAT application, the goal is to enumerate all the models of the propositional formula. 

Today, propositional satisfiability has gained a considerable audience with the advent of a new generation of solvers able to solve large instances encoding real-world problems. 
In addition to the traditional applications of SAT to hardware and software formal verification, this impressive progress led to increasing use of SAT technology to solve new real-world applications such as  planning, bioinformatics, cryptography.  In the majority of these applications, we are mainly interested in the decision problem and 
some of its optimisation variants (e.g. Max-SAT). Compared to other issues in SAT, the SAT model enumeration problem has received much less attention. Most of the recent proposed model enumeration approaches are built on the top of SAT solvers. Usually, these implementations are based on the  use of additional clauses, called blocking clauses, to avoid producing repeated models~\cite{McMillan02,Chauhan03usingsat,morgado05silva,Jin05efficientconflict}. 
Improvements have  been proposed to this blocking clause based enumeration solvers (e.g. \cite{Jin05efficientconflict,morgado05silva}). 
In particular, the authors in~\cite{morgado05silva} proposed several optimizations obtained through learning and simplification of  blocked clauses. 
However, these kind of approaches are clearly impractical. Indeed, in addition to clauses learned form conflicts, one might add an exponential number of blocked clauses in the worst case.  In \cite{MartinGebser07}, the authors elaborate an interesting approach for enumerating answer sets of a logic program (ASP), centered around First-UIP learning and backjumping. 

In~\cite{JabbourLSS14}, we proposed an approach based on a combination of a DPLL-like procedure with CDCL-based SAT solvers
in order to mainly avoid the limitation that concerns the space complexity
induced by blocking clauses addition. In this work, we focus on the SAT encoding of the problem of frequent itemset mining that we introduced in~\cite{JabbourSS13topk}. In such encoding, there is a one-to-one mapping between the models of the propositional formula and the set of interesting patterns  of the transaction database. Additionally, even for  condensed representation such as closed or maximal itemsets, the size of the output might be exponential in the worst case. 

The work presented in this paper is mainly motivated by this interesting SAT application to data mining and by the lack of efficient model enumerator. 


Our aim is to study through an extensive empirical evaluation, the effects on model enumeration of the main components of CDCL based SAT solvers including restarts, branching heuristics and clauses learning. 
 
\section{Background}

Let us first introduce the propositional satisfiability problem (SAT) and some necessary notations.  
We consider the  conjunctive normal form (CNF) representation for the propositional formulas. A {\it CNF formula}  $\Phi$  is a
conjunction ($\wedge$) of {clauses}, where a {\it clause} is a disjunction ($\vee$) of {literals}. 
A {\it literal} is a positive ($p$) or negated ($\neg{p}$) 
propositional variable.  The two literals $p$ and $\neg{p}$ are called {\it complementary}. A CNF formula can also be seen as a set of clauses, and a clause as a set of literals. 
Let us mention that any propositional formula can be translated to CNF using linear Tseitin's encoding \cite{Tseitin68}. 
We denote by $Var(\Phi)$ the set of propositional variables occurring in $\Phi$.

A {\it Boolean interpretation} ${\cal B}$ of a propositional formula $\Phi$ is a function which  associates a value ${\cal B}(p )\in\{0, 1\}$ ($0$ corresponds to $false$ and $1$ to $true$)
to the propositional variables $p \in Var(\Phi)$. It is extended to CNF formulas as usual.
 A {\it model} of a formula $\Phi$ is a Boolean  interpretation ${\cal B}$ that  satisfies the  formula, i.e., ${\cal B}(\Phi)=1$. 
  We note ${\cal M}(\Phi)$ the set of models of $\Phi$. {\it SAT problem} consists in deciding if a given  formula 
admits a model or not. 

Let us informally describe the most important components of modern SAT solvers. They are based on a reincarnation of the historical Davis, Putnam, Logemann and Loveland procedure, commonly called DPLL \cite{Davis62}. It performs a backtrack search; selecting at each level of the search tree, a decision variable which is set to a Boolean value. This assignment is followed by an inference step that deduces and propagates some forced unit literal assignments. This is recorded in the implication graph, a central data-structure, which encodes the decision literals   together with there implications. This branching process is repeated until finding a model or a conflict. In the first case, the formula is answered satisfiable, and the model is reported, whereas in the second case, a conflict clause (called learnt clause) is generated by resolution following a bottom-up traversal of the implication graph \cite{Marques-Silva96,Zhang01}. The learning or conflict analysis process stops when a conflict clause containing only one literal from the current decision level is generated. Such a conflict clause asserts that the unique literal with the current level (called asserting literal)  is implied  at a previous level, called assertion level,  identified as the maximum level of the other literals of the clause. The solver backtracks to the assertion level and assigns that asserting literal to \textit{true}. When an empty conflict clause is generated, the literal is implied at level 0, and the original formula can be reported unsatisfiable. In addition to this basic scheme, modern SAT solvers use other components such as activity based heuristics and restart policies. An extensive overview can be found in \cite{HandbookSAT}.

\section{Frequent Itemset Mining}


Formally, we define the problem of mining frequent itemsets (FMI for short) in the following way. 
Let $\Omega$ be a finite set of {\it items}. 
A {\it transaction} is defined as a couple $(tid,I)$ where $tid$ is the {\it transaction identifier} and 
$I$ is an {\it itemset}, i.e., $I\subseteq {\Omega}$. A {\it transaction database}  is a finite set of transactions 
where the attribute $tid$ refers to a unique itemset. 
We say that a transaction $(tid, I)$ {\it supports} an itemset $J$ if  $J\subseteq I$.\\
~\\
The {\it cover} of an itemset $I$ in a transaction database $\cal D$ is the set of
transactions  in   $\cal D$ supporting $I$: 
${\cal C}(I,{\cal D})=\{(tid,J)\in{\cal D}\mid  I\subseteq J \}$.
The {\it support} of an itemset $I$ in $\cal D$ is defined as the size of its cover:
${\cal S}(I,{\cal D})=\mid {\cal C}(I,{\cal D})\mid$.

\begin{table}
\begin{center}
\begin{tabular}{cccc}
\hline
Tid & Itemset \\
\hline
&&&\\
1 &  $A,B,C,D$ \\
2 & $A,B,E,F$ \\	
3 &  $A,B,C$ \\
4 & $A,C,D,F$ \\
5 & $G$ \\
6 & $D$\\\cal B
7 & $D,G$ \\
\hline
\end{tabular}
\end{center}
\caption{A transaction database $\cal D$}
\label{table1}
\end{table}
Let $\cal D$ be a transaction database over $\Omega$ and $n$ a minimum support threshold.
The {\it frequent itemset mining problem} consists in computing the following set:
$${\cal FIM}({\cal D},n)=\{I\subseteq {\Omega}\mid {\cal S}(I,{\cal D})\geq n\}. $$
\begin{definition}[Closed Frequent  Itemset]
Let $\cal D$ be a transaction database (over $\Omega$) and $I$ an itemset ($I\subseteq{\Omega}$)
such that ${\cal S}(I,{\cal D})\geq 1$.
The itemset $I$ is closed if, for all itemset $J\subseteq {\Omega}$ with $I\subset J$, we have that 
${\cal S}(J,{\cal D}) < {\cal S}(I,{\cal D})$.
\end{definition}
One can  see that all the elements of ${\cal FIM}({\cal D},n)$ can be obtained from the closed 
itemsets by computing their subsets. Enumerating all closed itemsets
allows us to reduce the size of the output. We denote by ${\cal CFIM}({\cal D},n)$ the subset of all closed itemsets in ${\cal FIM}({\cal D},n)$.

For instance, consider the transaction database described in Table~\ref{table1}. 
The closed frequent itemsets with the minimal support threshold equal to 2 are: 
${\cal CFIM}({\cal D}, 2)=\{A, D, G, AB, AC, AF, ABC,ACD\}$.

\section{A SAT Encoding of Frequent Itemset Mining}
\label{sec:enc}
In this section, we describe SAT encodings for itemset mining which are mainly based on the encodings proposed in~\cite{JabbourSS13topk}. 
 In order to do this, we fix, without loss of generality, a transaction database 
 ${\cal D}=\{(1,I_1), \ldots{}, (m, I_m)\}$ and a minimal support threshold $n$.
 
 The SAT encoding of itemset mining that we consider is based on the use of propositional variables representing the items and the transaction identifiers in $\cal D$.  
 More precisely, for each item $a$ (resp. transaction identifier $i$), we associate a propositional variable, denoted $p_a$ (resp. $q_i$).
 These propositional variables are used to capture all possible itemsets and their covers. Formally, given a model $\cal B$ of the considered encoding, 
 the candidate itemset is $\{a\in \Omega\mid {\cal B}(p_a)=1\}$ and its cover is $\{i\in \mathbb{N}\mid {\cal B}(q_i)=1\}$.
 
 The first propositional formula that we describe allows us to obtain the cover of the candidate itemset: 
 \begin{equation}
 \label{form:1}
 \bigwedge_{i=1}^m (\neg q_i \leftrightarrow \bigvee_{a\in \Omega\setminus I_i} p_a)
 \end{equation}
 This formula expresses that $q_i$ is true if and only if the candidate itemset is supported by the $i^{th}$ transaction. In other words, the candidate itemset is not supported by the $i^{th}$ transaction ($q_i$ is false), when there exists an item $a$ ($p_a$ is true)  that does not belong to the transaction ($a\in \Omega\setminus I_i$).
 
 The following propositional formula allows us to consider the itemsets having a support greater than or equal to 
 the minimal support threshold: 
  \begin{equation}
   \label{form:2}
\sum_{i=1}^m q_i \geq n
 \end{equation}
This formula corresponds to 0/1 linear inequalities, usually called cardinality  constraints.  
The first linear encoding of general 0/1 linear inequalities to CNF have been proposed by J. P. Warners in~\cite{Warners96}. 
Several authors have addressed the issue of finding an efficient encoding of cardinality (e.g. \cite{Sinz05,SilvaL07,Asin11}) as a CNF formula. 
Efficiency refers to both the compactness of the representation (size of the CNF formula)  and to the ability to achieve the same level of constraint propagation (generalized arc consistency) 
on the CNF formula.\\

We use ${\cal E}_{FIM}({\cal D},n)$ to denote the encoding corresponding to the conjunction of the two formul\ae{} ($\ref{form:1}$) and ($\ref{form:2}$).
Then, we have the following property:
$\cal B$ is a model of ${\cal E}_{FIM}({\cal D},n)$ iff $I=\{a\in\Omega\mid {\cal B}(p_a)=1\}$ is a frequent itemset where ${\cal C}(I,{\cal D})= \{i\in\mathbb{N}\mid {\cal B}(q_i)=1 \}$.

We now describe the propositional formula allowing to force the candidate itemset to be closed: 
\begin{equation}
\label{form:3}
\bigwedge_{a\in {\Omega}} (\bigwedge_{i=1}^{m} q_{i}\rightarrow a\in I_i)\rightarrow   p_{a}
\end{equation} 
This formula means that if we have ${\cal S}(I, {\cal D})={\cal S}(I\cup\{a\}, {\cal D})$ then $a\in I$ holds. 
This condition is necessary and sufficient to force the candidate itemset to be closed. 
Let us note that the expressions  of the form $a\in I_i$ correspond to constants, i.e.,  $a\in I_i$ corresponds to $\top$ if the item $a$ is in $I_i$, to $\bot$ otherwise.

Note that the formula ($\ref{form:3}$) can be simply  reformulated as a conjunction of clauses as follows: 
\begin{equation}
\label{form:32}
\bigwedge_{a\in {\Omega}} ( (\bigvee_{ 1\leq i\leq m, a\not\in I_i,} q_{i})\vee  p_{a})
\end{equation} 
This reformulation is obtained using the equivalence $A\rightarrow B\equiv\neg A\vee B$.

We use ${\cal E}_{CFIM}({\cal D},n)$ to denote the encoding corresponding to the conjunction of the formul\ae{}  ($\ref{form:1}$), ($\ref{form:2}$) and ($\ref{form:32}$).
Then, we have the following property:
$\cal B$ is a model of ${\cal E}_{CFIM}({\cal D},n)$ iff $I=\{a\in\Omega\mid {\cal B}(p_a)=1\}$ is a closed frequent itemset where ${\cal C}(I,{\cal D})= \{i\in\mathbb{N}\mid {\cal B}(q_i)=1 \}$.

\begin{example}
Let us consider the transaction database of Table \ref{table1}. The Problem encoding the enumeration of frequent closed itemsets with a threshold 4 can be written as: \\

\begin{tabular}{ll}
 & $ \{ \neg q_1 \leftrightarrow (p_{E} \vee p_{F} \vee p_{G})$,\\
&$\neg q_2 \leftrightarrow (p_{C} \vee p_{D} \vee p_{G})$,\\
&$\neg q_3 \leftrightarrow (p_{D} \vee p_{E} \vee p_{F} \vee p_{G})$,\\
&$\neg q_4 \leftrightarrow (p_{B} \vee p_{E} \vee p_{G})$,\\
&$\neg q_5 \leftrightarrow (p_{A} \vee p_{B} \vee p_{C} \vee p_{D} \vee p_{E} \vee p_{F})$,\\
&$\neg q_6 \leftrightarrow (p_{A} \vee p_{B} \vee p_{C}  \vee p_{E} \vee p_{F} \vee p_{G})$,\\
&$\neg q_5 \leftrightarrow (p_{A} \vee p_{B} \vee p_{C} \vee p_{E} \vee p_{F}),$\\
&($q_5 \vee q_6 \vee q_7 \vee p_{A}$), \\
&($q_4 \vee q_5 \vee q_6 \vee q_7 \vee p_{B}$), \\
&($q_2 \vee q_5 \vee q_6 \vee q_7 \vee p_{C}$), \\
&($q_2 \vee q_3 \vee q_5  \vee p_{D}$), \\
&($q_2 \vee q_3 \vee q_4 \vee q_5 \vee q_6 \vee q_7 \vee p_{E}$), \\
&($q_1 \vee q_3 \vee q_4 \vee q_5 \vee q_6 \vee q_7 \vee p_{F}$),\\
&($q_1 \vee q_2 \vee q_3 \vee q_4 \vee q_6 \vee  p_{G}$),\\
& $q_1 + q_2 + q_3 + q_4 +q_5 + q_6 + q_7 \ge 4 \}$
\end{tabular}
\end{example}

\section{Enumerating all Models of CNF Formulae} 
\label{sec:enum}
A naive way to extend modern SAT solvers for the problem of enumerating all models of a CNF formula consists in adding a blocking clause to prevent the search to return the same model again. This approach is used in the majority of the model enumeration methods in the literature. 
The main limitation of this approach concerns the space complexity, since the number of blocking clauses may be exponential in the worst case. 
Indeed, in addition to the clauses learned at each conflict by the CDCL-based SAT solver, the number of added blocking clauses is very important on problems with a huge number of models.
This explains why it is necessary to design methods avoiding the need to keep all blocking clauses. 
It is particularly the case for encodings of data mining tasks where the number of interesting patterns is often significant, even when using condensed representations such as closed patterns.

Our main aim is to experimentally study the effects of each component of modern SAT solvers on the efficiency of the model enumeration, in the case of the encoding described in Section~\ref{sec:enc}. 
As the number of  frequent closed itemsets is usually huge, this results in a huge number of models for the considered encoding.
Consequently, it is not suitable to store the found models using blocking clauses during the enumeration process. 

We proceed by removing incrementally some components of modern SAT solvers in order to evaluate their effects on the efficiency of model enumeration. The first removed component is  the restart policy.  
Indeed, we inhibit the restart in order to allow solvers to avoid the use of blocking clauses. Thus, our procedure performs a simple backtracking at each found model.
The second removed component is that of  clause learning, which leads to a DPLL-like procedure.  
Considering a DPLL-like procedure,  we pursue our analysis by considering the  branching heuristics.  Indeed, our goal is to find the heuristics suitable to the considered SAT encoding.  
To this end, we consider three branching  heuristics. We first study the performance of the well-known VSIDS (Variable State Independent, Decaying Sum) branching heuristic. 
In this case, at each conflict an analysis is only performed to weight variables (no learnt clause is added).
The second considered branching heuristic is based on the maximum number of occurrences of the the variables. The third one consists in selecting the variables randomly. 


\section{Experiments}

We carried out an experimental evaluation to analyze the effects of adding blocking clauses, adding learned clauses and branching heuristics. To this end, 
we implemented a DPLL-like procedure, denoted {\tt DPLL-Enum}, without adding blocking and learned clauses. We also implemented a procedure on the top 
of the state-of-the-art CDCL SAT solver MiniSAT 2.2, denoted {\tt CDCL-Enum}. In this procedure, each time a model is found, we add a no-good and perform a restart. 
 We considered a variety of datasets taken from the {\tt FIMI}\footnote{FIMI: http://fimi.ua.ac.be/data/} and {\tt CP4IM}\footnote{CP4IM: http://dtai.cs.kuleuven.be/CP4IM/datasets/} repositories.  
 All the experiments were done on Intel Xeon quad-core machines with 32GB of RAM running at 2.66 Ghz. For each instance, we used a timeout of 15 minutes of CPU time.


In our experiments, we compare the performances of {\tt CDCL-Enum} to three variants of {\tt DPLL-Enum}, with different branching heuristics, in enumerating all the models corresponding to the closed frequent itemsets. The considered variants of {\tt DPLL-Enum} are the following: 
\begin{itemize}
\item {\tt DPLL-Enum+VSIDS}:  {\tt DPLL-Enum} with the VSIDS branching heuristic; 
\item {\tt DPLL-Enum+JW}:  {\tt DPLL-Enum} with a branching heuristic based on the maximum number of occurrences of the variables \cite{Jeroslow90}; 
\item {\tt DPLL-Enum+RAND}: {\tt DPLL-Enum} with a random variable selection. 
\end{itemize}

\begin{figure}[htbp]
\centering
\begin{tabular}{cc}
  \includegraphics[width=5cm]{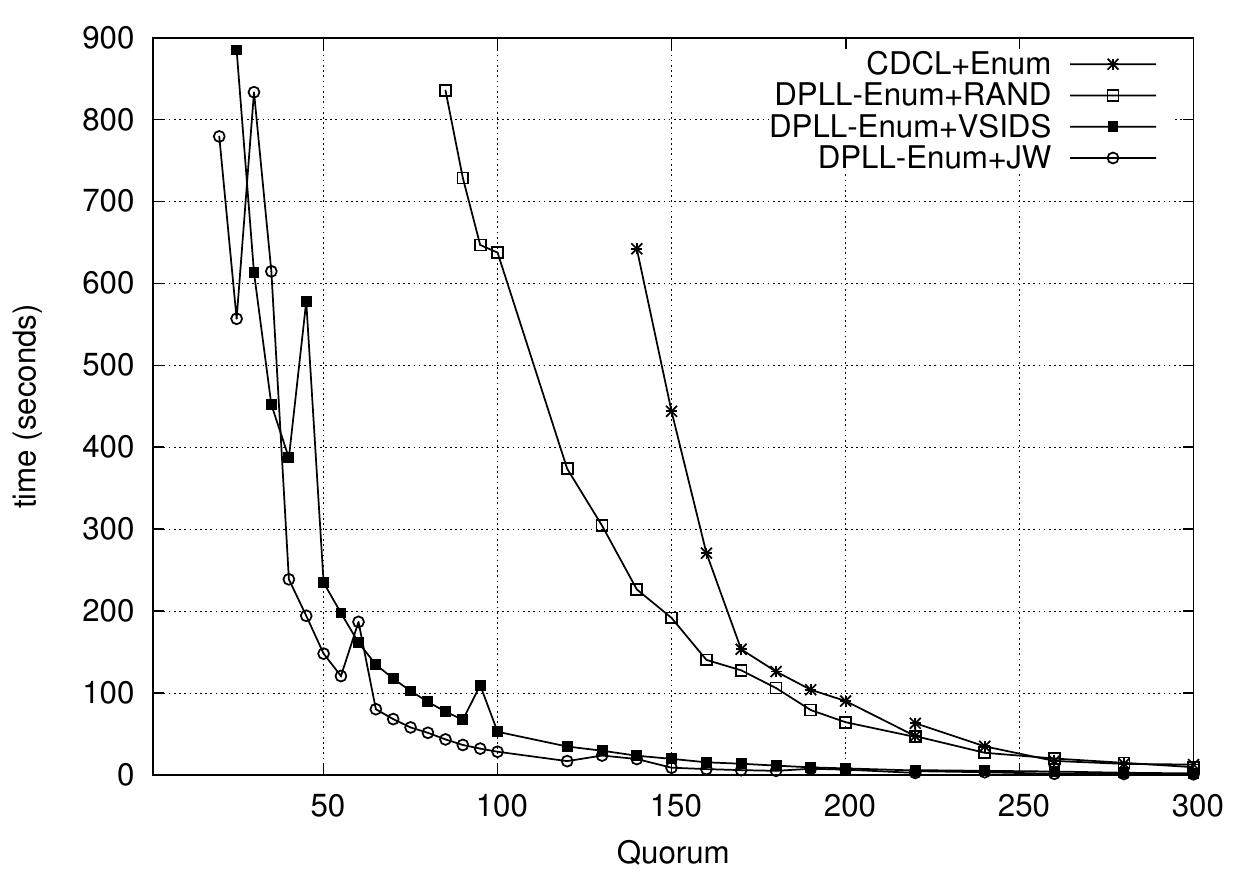}   & \includegraphics[width=5cm]{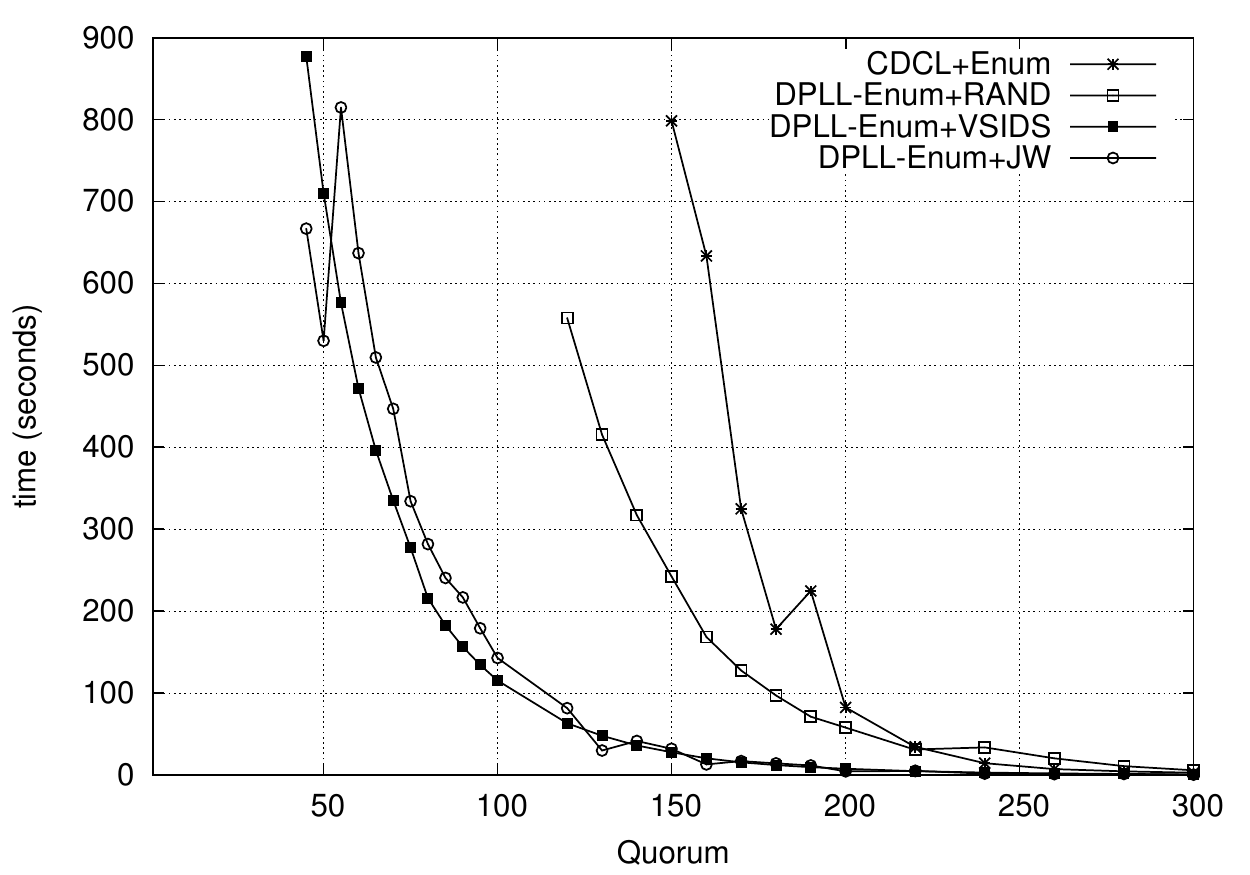}    \\
  german-credit & australian-credit \\
   \includegraphics[width=5cm]{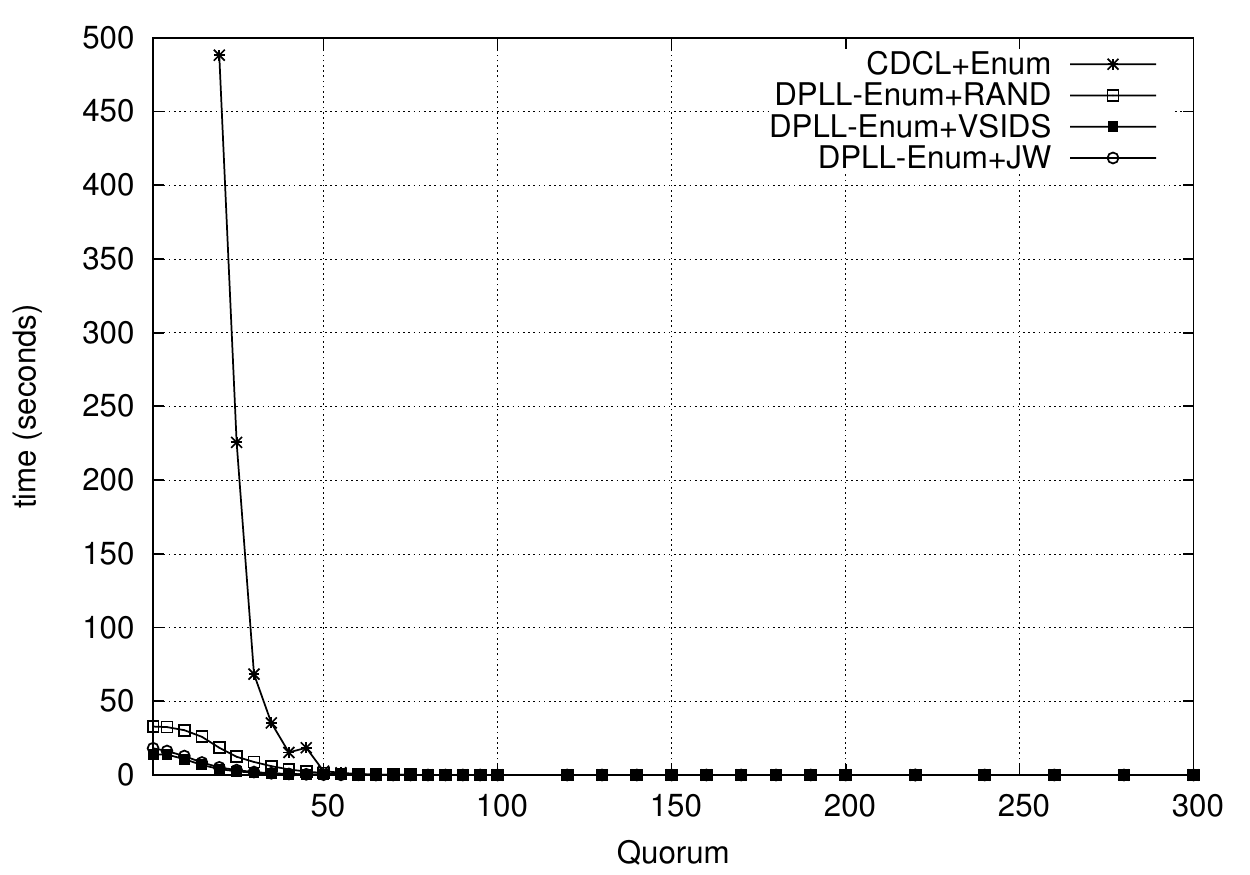}   & \includegraphics[width=5cm]{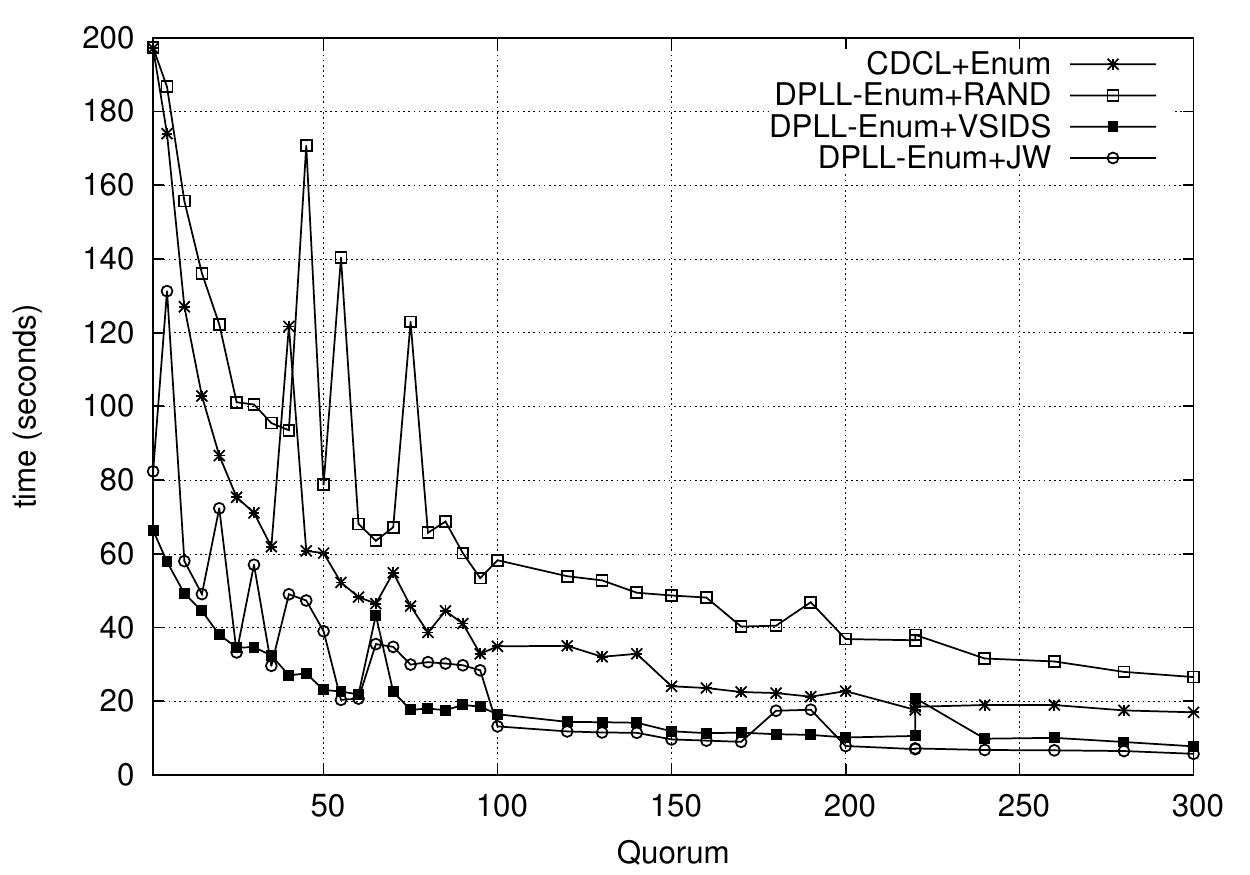}    \\
  hepatitis.pdf & mushroom \\
    \includegraphics[width=5cm]{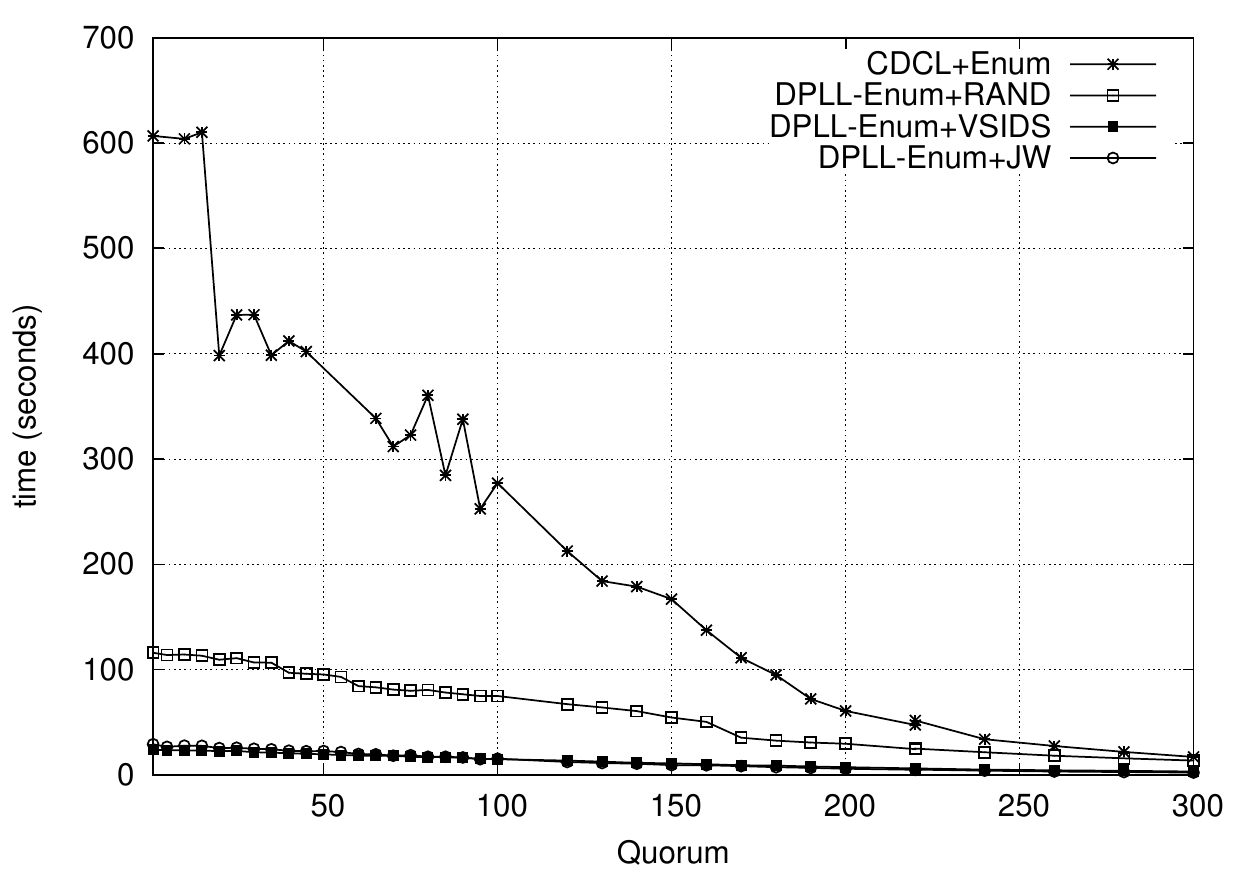}   & \includegraphics[width=5cm]{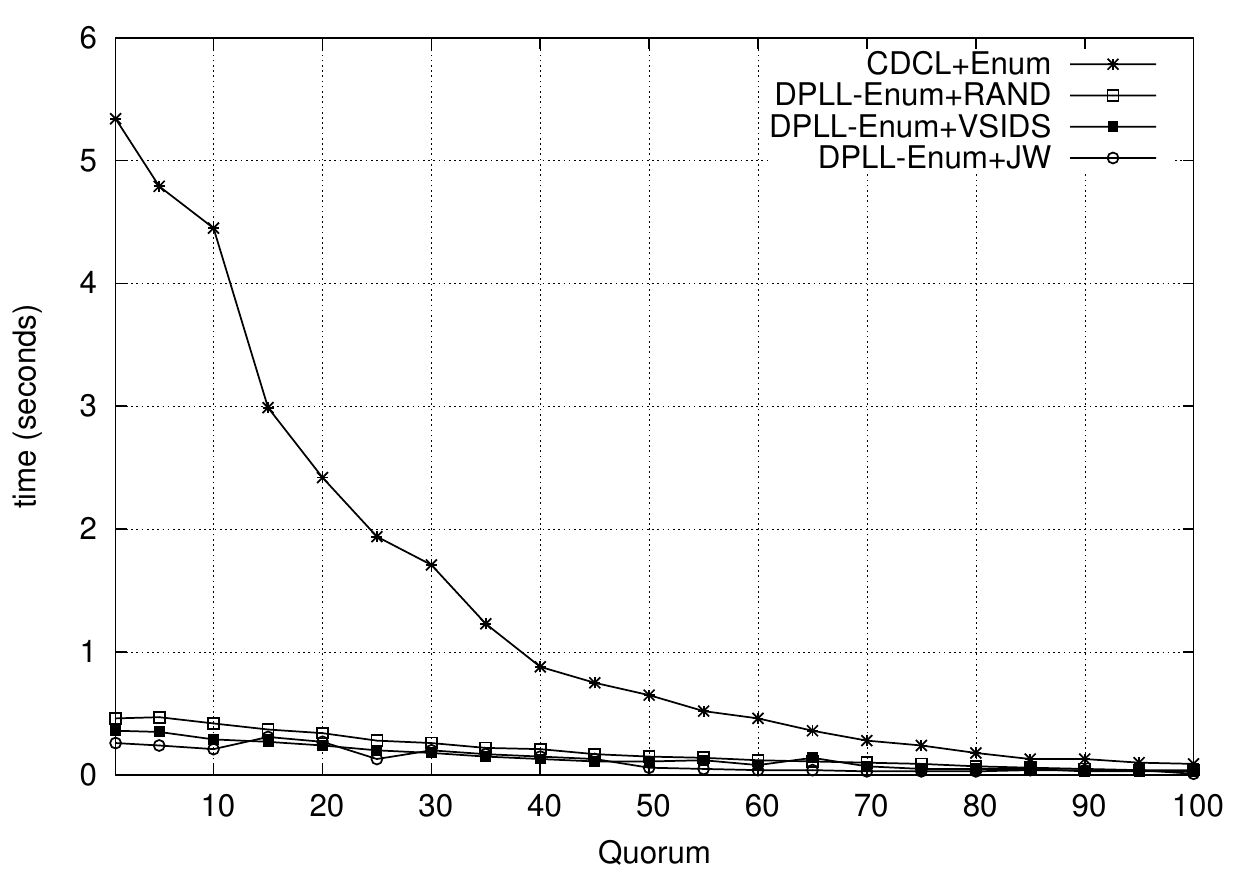}    \\
  anneal & primary-tumor \\  
      \includegraphics[width=5cm]{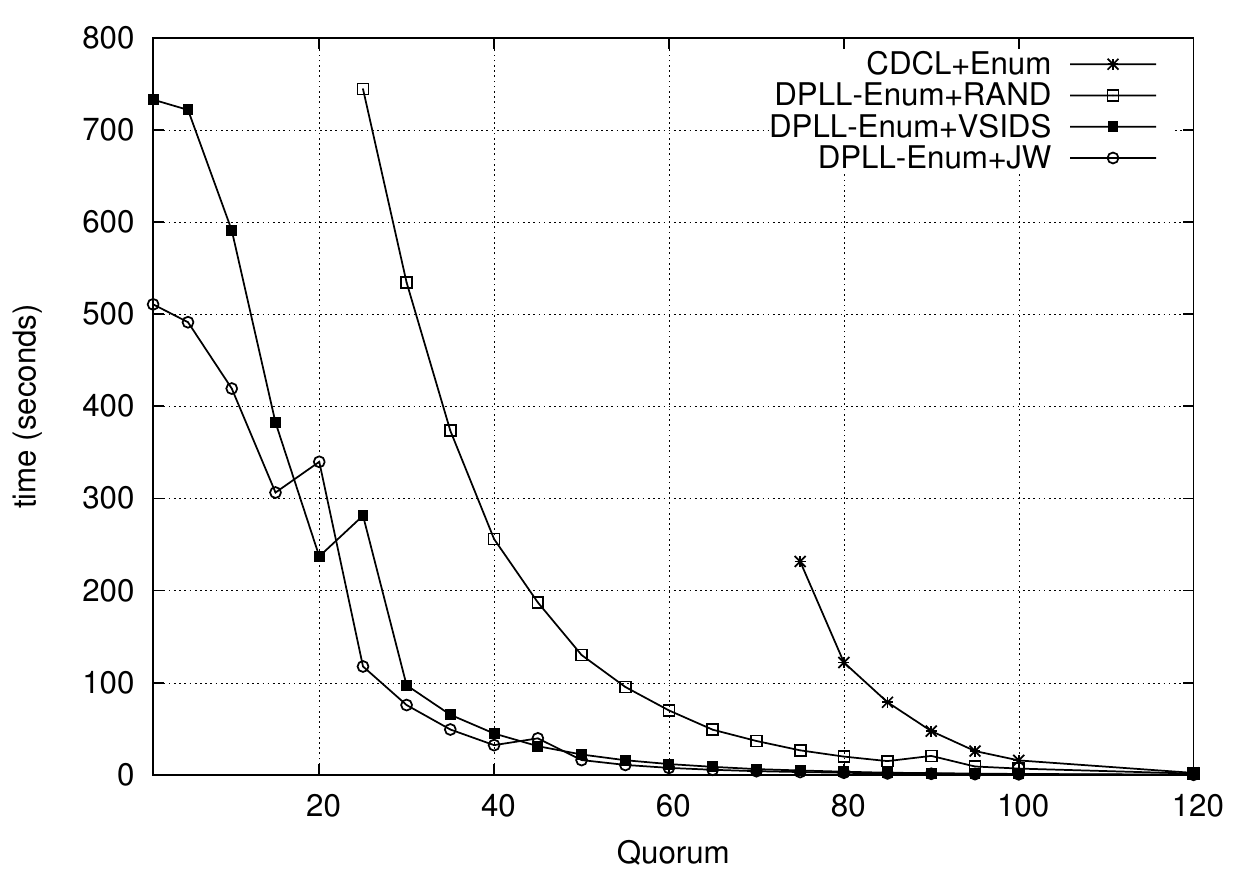}   & \includegraphics[width=5cm]{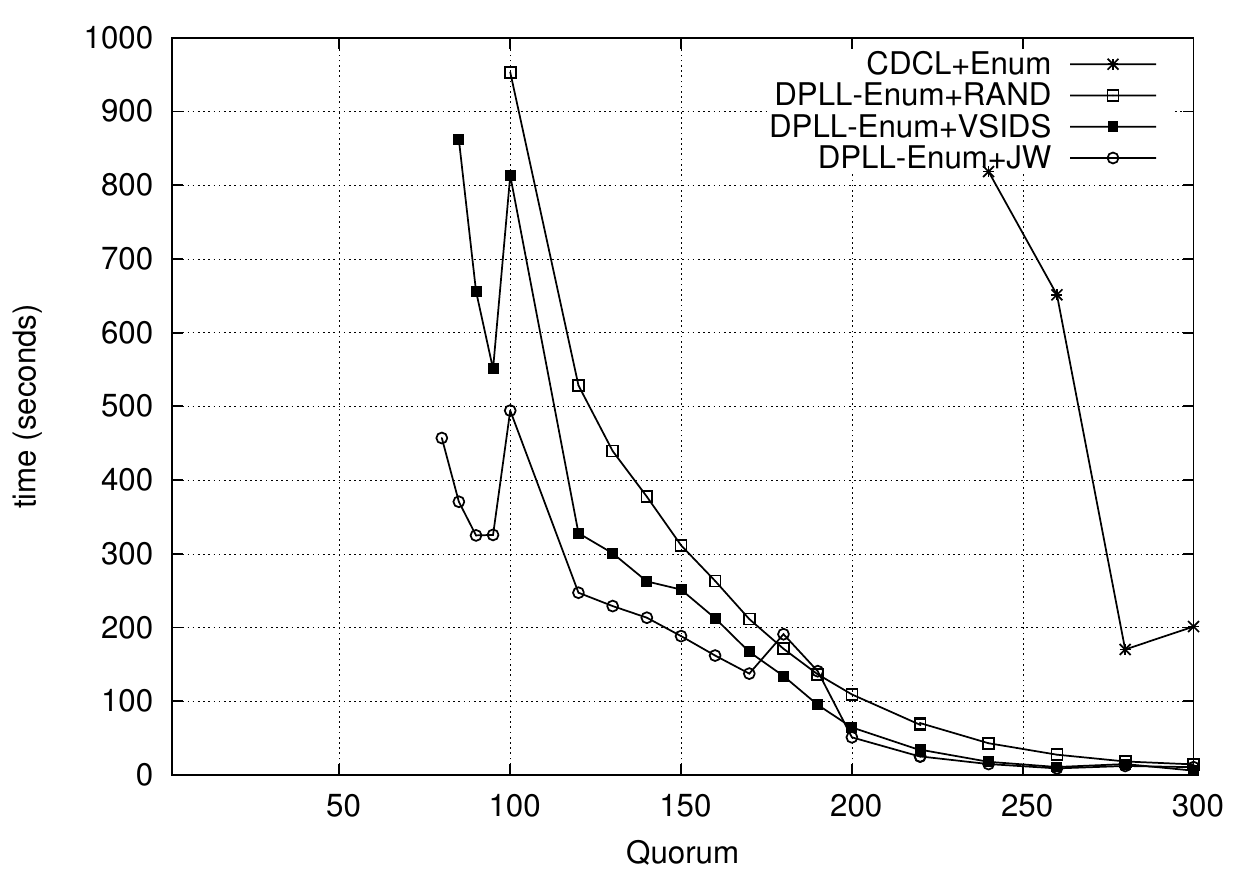}    \\
  heart-cleveland.pdf & splice-1 \\   
        \includegraphics[width=5cm]{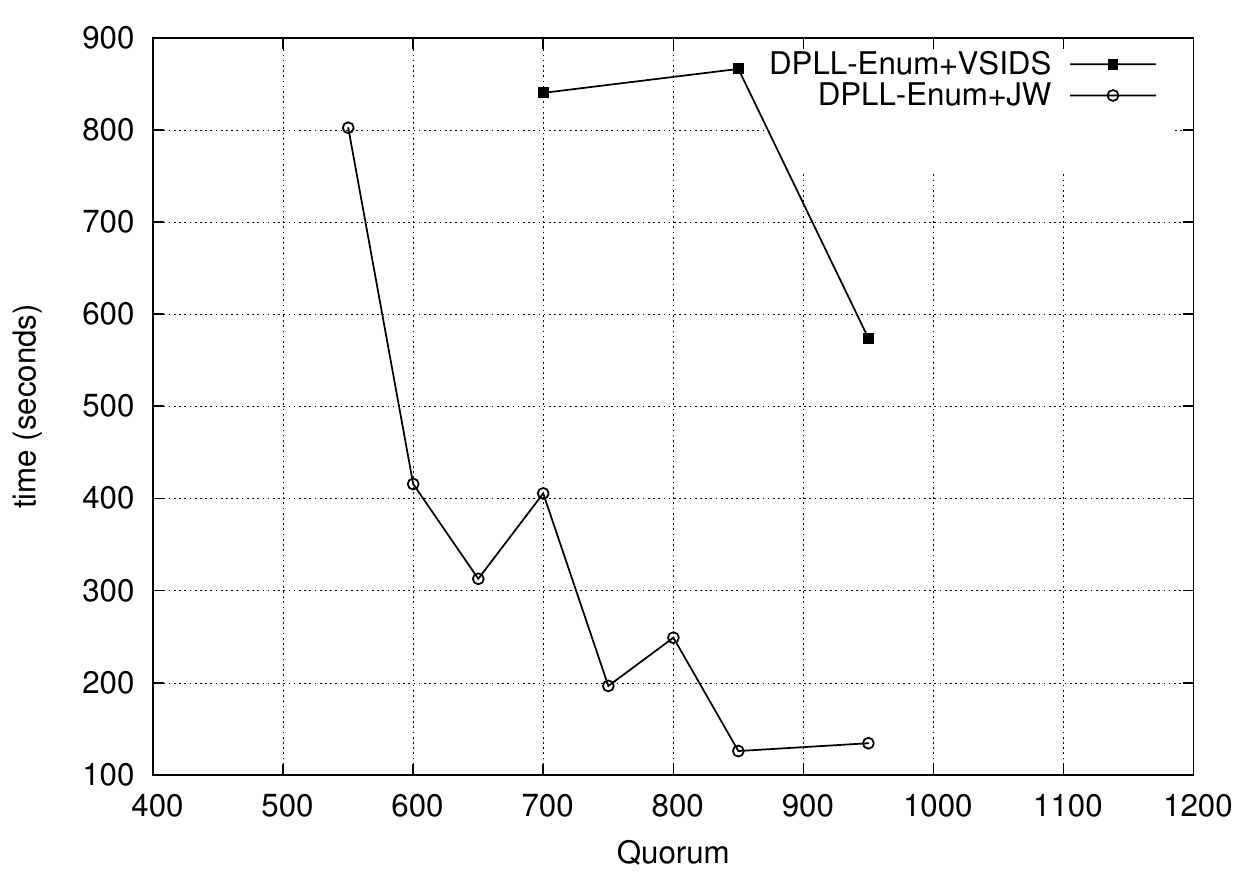}   & \includegraphics[width=5cm]{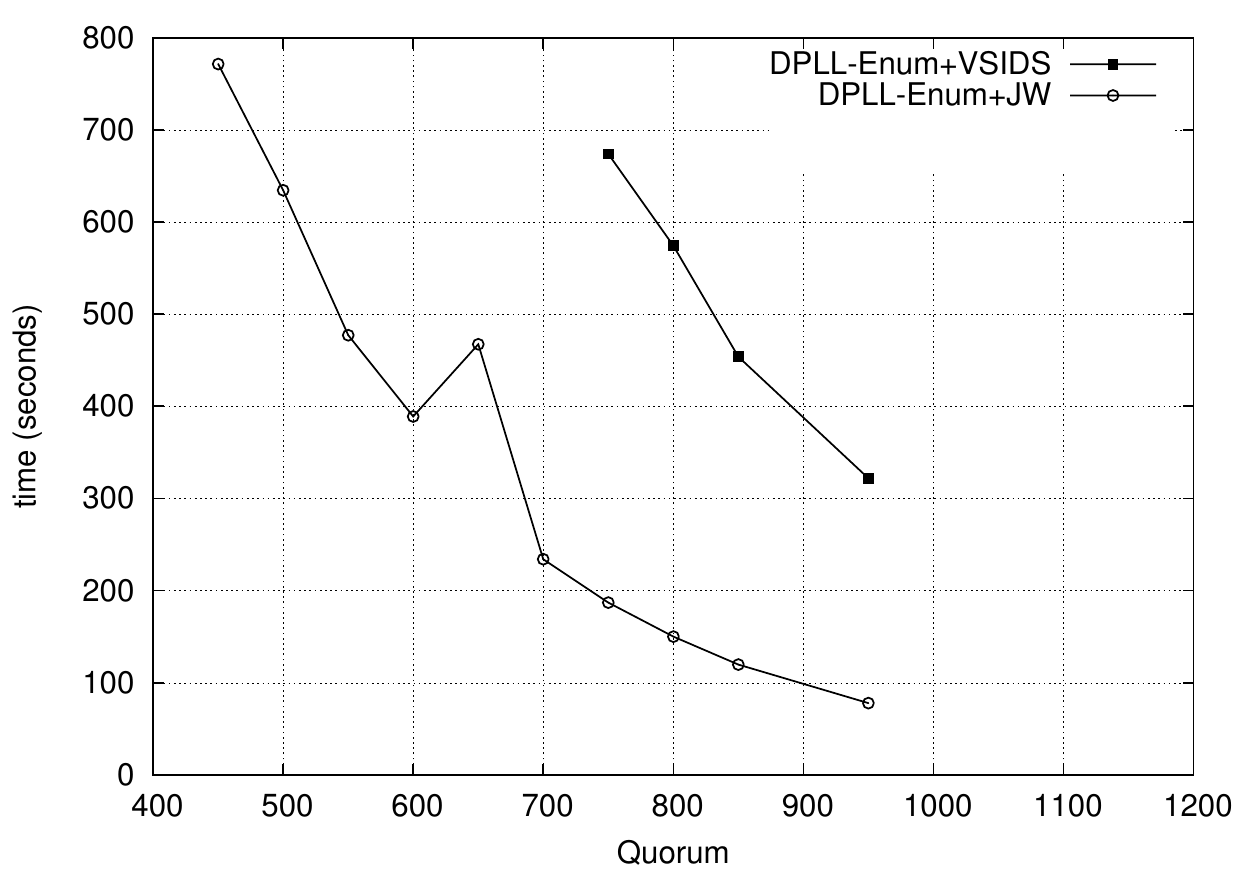}    \\        
  chess & kr-vs-kp \\      
  \end{tabular}
\caption{Frequent Closed Itemsets: CDCL vs DPLL-Like enumeration} 
\label{fig:scatter}
\end{figure}

Our comparison is depicted by the the cactus plots of Figure \ref{fig:scatter}.  Each dots $(x, y)$ represents an instance  with a fixed minimal support threshold $n$. Each cactus plot represents an instance and the evolution of CPU time needed to enumerate all models with the different algorithms while varying the quorum. For each instance, we tested different values of $n$.   The x-axis (respectively y-axis) represents the CPU time (in seconds) needed for the enumeration of all closed frequent itemsets. 



Unsurprisingly, the DPLL-like procedures outperform {\tt CDCL-Enum} on the majority of instances. 
This shows that a DPLL  based approach is more suitable for SAT-based itemset mining. 
Part of explanation lies in the significant number of models. Furthermore, {\tt DPLL-Enum+RAND} 
is clearly less efficient than  {\tt DPLL-Enum+VSIDS} and {\tt DPLL-Enum+JW}, which shows that the branching heuristic 
plays a key role in model enumeration algorithms. Moreover, our experiments show that {\tt DPLL-Enum+JW} is better than 
{\tt DPLL-Enum+VSIDS}, even if {\tt DPLL-Enum+VSIDS}  compete with the procedure {\tt DPLL-Enum+JW} on datasets such as {\tt anneal} and {\tt mushroom}.
Indeed, {\tt DPLL-Enum+JW} clearly outperforms {\tt DPLL-Enum+VSIDS} on several datasets, such as {\tt chess}, {\tt kr-vs-kp} and {\tt splice-1}. Note that for theses two data, the solvers {\tt DPLL-Enum+RAND} and {\tt CDCL-Enum} are not able to enumerate completely the set of all models of all considered quorums.

 As a summary, our experimental evaluation suggests that a DPLL-like procedure is a more suitable approach when the number of models of a propositional formula is significant. 
 It also suggests that the branching heuristic is a key point in such a procedure to improve the performance.

\section{Conclusion}
In this paper, we investigated the impact of modern SAT solvers on the problem of enumerating all the models of CNF formulas encoding frequent closed itemsets mining problem. Our goal is to measure the impact of the classical components of CDCL-based SAT solvers on the efficiency of model enumeration. Our results suggest that on formula with a huge number of models, SAT solvers must be adapted to efficiently enumerate all the models. We showed that the simple DPLL solver augmented with the classical Jeroslow-Wang heuristic achieve better performance.  


As a future work, we plan to pursue our investigation in order to find the best heuristics for enumerating models encoding data mining problems. Finding how to efficiently integrate clauses learning for model enumeration is another interesting issue.


\bibliographystyle{plain}
\bibliography{satBib}
\end{document}